# An Empirical Study of Incremental Learning in Neural Network with Noisy Training Set

Atrayee Chatterjee[1,*], Shovik Ganguly[2,†], Debasmita Bhoumik[3,#], Ritajit Majumdar[4,$]

[1]Asutosh College, University of Calcutta
[2]Lexmark International India Pvt. Ltd.
[3,4]Advanced Computing and Microelectronics Unit, Indian Statistical Institute, Kolkata
Email: [*]atrayeechatterjeeiii@gmail.com, [†]gangulyshovik@gmail.com,
[#]debasmita.ria21@gmail.com, [$]majumdar.ritajit@gmail.com

**Abstract.** The notion of incremental learning is to train an ANN algorithm in stages, as and when newer training data arrives. Incremental learning is becoming widespread in recent times with the advent of deep learning. Noise in the training data reduces the accuracy of the algorithm. In this paper, we make an empirical study of the effect of noise in the training phase. We numerically show that the accuracy of the algorithm is dependent more on the location of the error than the percentage of error. Using Perceptron, Feed Forward Neural Network and Radial Basis Function Neural Network, we show that for the same percentage of error, the accuracy of the algorithm significantly varies with the location of error. Furthermore, our results show that the dependence of the accuracy with the location of error is independent of the algorithm. However, the slope of the degradation curve decreases with more sophisticated algorithms.

## 1 Introduction

Machine learning (ML) algorithms aim to train a computer to make decisions [1]. These algorithms are used in various fields such as image processing [2], [3], object recognition [4], handwriting recognition [5], natural language processing [6] and even quantum computing [7]. There are two major techniques for decision making used by ML algorithms - (i) Supervised learning where the algorithm is initially trained with a set of labelled data [8], and (ii) Unsupervised learning where the algorithm looks for patterns and similarities and tries to group similar data in the same cluster without any prior learning phase [9]. Although these two are the major types of learning algorithms, other forms of learning algorithms such as semi-supervised learning [10], reinforcement learning [11] are also widely studied.

Artificial Neural Network (ANN) algorithms are supervised algorithms which mimic the working principles of neurons in the human brain. The most basic ANN algorithm is the single-layer Perceptron. However, more sophisticated algorithms use complicated functions for decision making, and take the error of the output into consideration in order to update the weights. Although these algorithms assume a single initial training phase, in real world data often arrives in batches, which requires the training to be performed in multiple stages. The model is first trained with the initial set of training data, and when more training data is available, the model is trained further. Such a model of training is called Incremental Learning [12].

In this paper, we study incremental learning in ANN where the training dataset may be noisy, i.e., some of the training data have incorrect label. It is obvious that if the training dataset is erroneous, the training will be less effective and the ANN will be less accurate in its prediction. In this paper, we show, however, that not only the number of erroneous data, but also the location of the error in incremental learning plays a vital role in the performance of the algorithm. The accuracy of an algorithm varies even when the percentage of error is the same, but the errors are concentrated in different locations of the training set.

In this paper, we show numerically by using three ANN algorithms (Perceptron [13], Feedforward Neural Network (FFN) [14], [15] and Radial Basis Functions Neural Network (RBF) [15]) that (i) for the same percentage of erroneous data, the location of error clusters can significantly alter the performance of the algorithm, (ii) the performance degradation is independent of the number of features per data and (iii) although more sophisticated algorithms are more robust to errors, the degradation in performance due to concentrated error has a similar nature for all algorithms.

The rest of the paper is organized as follows - in Section 2, we give a brief description of the three ANN algorithms used. In Sections 3 and 4, we show the performance of the algorithms for two steps and three steps incremental learning. We conclude in Section 5.

## 2   Brief review of the ANN algorithms used

An $m$-class classification problem [16], where $C_1, C_2, ..., C_m$ are the object classes, is associated with $k$ training samples and $n$ testing samples. Each training sample is a vector $(s_i, C_{si})$, where $C_{si}$ is the designated class of the sample $s_i$. After training the algorithm with this set of training data, for each testing sample $t_i$, such that $t_i$ belongs to class $C_j$, the algorithm is expected to produce

$$Prob(t_i \in C_j) > Prob(t_i \in C_l) \ \forall l \neq j. \tag{1}$$

The algorithm is said to have made an error in the prediction if for some testing sample $t_p \in C_p$, it produces $Prob(t_p \in C_p) < Prob(t_q \in C_q)$ for some $q \neq p$. The objective of learning is to minimize the number of errors. In the following part of this subsection we briefly discuss the working principles of the three ANN algorithms (Perceptron, Feed Forward Neural Network and RBF) used in this paper. Our motivation behind using these three algorithms is to show that although sophistication of the algorithm enhances the robustness to training errors, the performance loss due to the location of error concentration remains invariant under the type of algorithm used.

**Perceptron Learning Algorithm.** Perceptron is one of the simplest ANN algorithms. In this algorithm, an input is a vector $(x_1, x_2, ..., x_n)$, where each $x_i$ is called a feature and associated with each feature is a weight $w_i$. For a 2-class classification problem, the output

$$y = \begin{cases} 1 & if \ \sum_{i=1}^{n} w_i x_i \geq \theta \\ 0 & otherwise \end{cases} \quad (2)$$

where $\theta$ is a threshold value. The weights $w_i$ are initialized randomly, and are modified during the training phase to match the class labels of each training sample. This algorithm can be easily modified for multi-class classification.

**Feedforward Neural Network Algorithm.** In FFN, the perceptrons are arranged in layers, with the first layer taking in inputs and the last layer producing outputs. The middle layers are called hidden layers. Each perceptron in one layer is connected to every perceptron on the next layer, but there is no interconnection among the perceptrons of the same layer. The information is constantly fed forward from one layer to the next. A single perceptron can classify points into two regions which are linearly separable, whereas by varying the number of layers, the number of input, output and hidden nodes one can classify points in arbitrary dimension into an arbitrary number of groups.

**Radial Basis Function Algorithm.** A drawback of Perceptron is that the activation function is linear, and hence fails to classify non-linearly separated data. The RBF Algorithm uses radial basis functions as activation functions. RBF is a real-valued function $\varphi$ whose value depends only on the distance from the origin $\varphi(x) = \varphi(\|x\|)$. Its characteristic feature is that the response decreases or increases monotonically with distance from a central point. A typical radial function is the Gaussian function. RBF transforms the input signal into a different form, which can then be fed to the ANN to get linear separability. RBF has an input layer, a single hidden layer, and an output layer. The sophistication of the activation function of this algorithm usually leads to better classification accuracy.

## 3 Performance of ANN in noisy two-step incremental learning

We have performed our study on the standard benchmark dataset IRIS [17], which contains 120 data samples, where each sample is a vector containing 4 features. 40 samples have been used t0 train each ANN algorithm, while the other 80 have been used to test their performance.

Error in training data implies that for a particular training sample $s_i \in C_i$, the class is incorrectly labelled as $C_j, j \neq i$. For this dataset, containing $2n$ training samples (here n = 20), each ANN algorithm is trained twice sequentially with the first $n$ and the last $n$ training samples. We have varied the errors in the training set from 0% to 50% by a gap of 10. Three scenarios are considered, where the erroneous data are (i) uniformly distributed in the entire training set, (ii) uniformly distributed in the first half of the training set, and (iii) uniformly distributed in the second half of the training set.

In Table 1 we show the accuracy obtained by Perceptron, FFN and RBF algorithms respectively as the error in the training sample is varied as discussed above.

**Table 1.** Accuracy of perceptron, FFN and RBF in the presence of clustered noise

| Error % | Error distributed uniformly in | | | | | | | | |
|---|---|---|---|---|---|---|---|---|---|
| | **Perceptron** | | | **FFN** | | | **RBF** | | |
| | entire dataset | first half | second half | entire dataset | first half | second half | entire dataset | first half | second half |
| 0 | 100 | 100 | 100 | 100 | 100 | 100 | 100 | 100 | 100 |
| 10 | 96.45 | 96.67 | 100 | 96.67 | 96.67 | 100 | 97.8 | 97 | 100 |
| 20 | 60 | 60 | 100 | 60 | 60 | 100 | 70.65 | 72 | 100 |
| 30 | 46.67 | 45.2 | 93.33 | 49 | 43.67 | 96.67 | 48.9 | 52.45 | 98.2 |
| 40 | 33 | 30 | 92.4 | 33 | 30 | 92.52 | 40.2 | 38 | 96.47 |
| 50 | 30.44 | 28.55 | 85.69 | 30 | 28.40 | 88.54 | 36 | 36 | 94.44 |

## 4  Performance of ANN in noisy three-step incremental learning

We have performed this study on the WINE dataset [17], which contains 150 data samples, where each sample is a vector containing 13 features. The IRIS dataset chosen in the previous two step experiment does not contain enough samples to effectively divide into three training sets. As such we chose the WINE dataset for this experiment. For this dataset, the ANN algorithms have been trained using 60 samples and the other 90 have been used to test their performance. The motivation to use these two different datasets (IRIS and WINE) is to show that the performance degradation due to error clusters is independent of the number of features per sample. For the WINE dataset containing $3n$ training samples, each ANN algorithm is trained thrice sequentially with the first $n$, second $n$ and the last $n$ training samples. We have varied the errors in the training set from 0% to 40% by a gap of 10. Three scenarios are considered in each case, where the erroneous data are (i) uniformly distributed in the entire training set, (ii) uniformly distributed in the first 20 entries, (iii) uniformly distributed in the second 20 entries, and (iv) uniformly distributed in the last 20 entries of the training set.

In Fig 1, we show the graph of the accuracy of the three algorithms for incremental training. The first row shows the performance for two-step training, and the second row shows the same for three-step training. The graphs readily show that the degradation in performance is heavily dependent on the location of error. Performance of all the algorithms, when errors are distributed uniformly or are clustered in the first training set, is almost the same. However, as the errors move to later training sets, the performance of the algorithm increases significantly. Moreover, although the performance of RBF is better than FFN, which in its turn is better than perceptron, the nature of degradation remains similar for all the algorithms, irrespective of its sophistication.

For the three step learning, we have also studied the accuracy of the said algorithms when the error is uniformly distributed in two of the three steps. The graph of the accuracy in this scenario is shown in Fig 2.

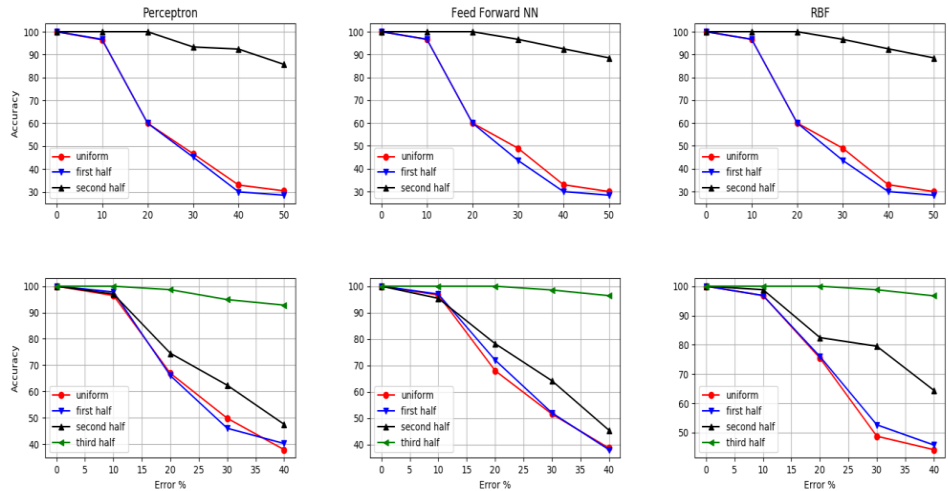

**Fig 1. Accuracy of Perceptron, FFN & RBF for two and three steps incremental learning**

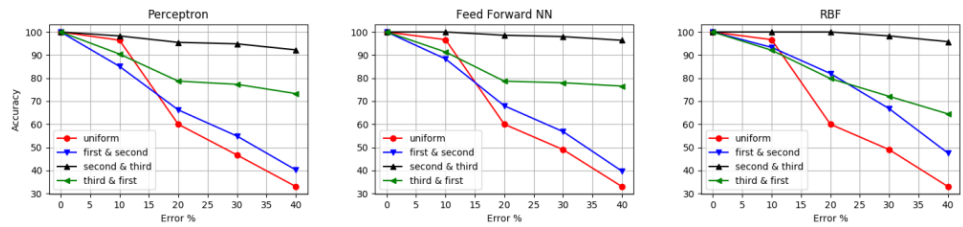

**Fig 2. Accuracy of errors uniformly distributed in two of three steps**

## 5  Conclusion

In this paper we have numerically studied the accuracy of Perceptron, FFN and RBF for two steps and three steps incremental learning in the presence of noisy dataset. We show that the accuracy of the algorithms depends not only on the percentage of error on the training set, but also on the location of the error concentration. In fact, for the same percentage of error, the location plays a significant role in the accuracy of the algorithms. Furthermore, results obtained from the most basic ANN (Perceptron) and a more sophisticated ANN (FFN, RBF) shows that although sophistication makes the algorithm more robust to errors, the nature of the performance degradation due to location of error has similar effect on all the algorithms.